\documentclass[10pt,twocolumn,letterpaper]{article}

\usepackage{cvpr}
\usepackage{times}
\usepackage{epsfig}
\usepackage{graphicx}
\usepackage{amsmath}
\usepackage{amssymb}
\usepackage{soul}
\usepackage{dblfloatfix}
\usepackage{subfig}
\usepackage{subcaption}
\usepackage[font=small]{caption}

\begin{document}

\title{FDWST: Fingerphoto Deblurring using Wavelet Style Transfer}
\author{
David Keaton, 
Amol S. Joshi, 
Jeremy Dawson, 
Nasser M. Nasrabadi\vspace{3pt}\\
West Virginia University\\
{\tt\small \{dck0013, asj00003\}@mix.wvu.edu, \{jeremy.dawson, nasser.nasrabadi\}@mail.wvu.edu}
}

\maketitle
\thispagestyle{empty}

\begin{abstract}
The challenge of deblurring fingerphoto images, or generating a sharp fingerphoto from a given blurry one, is a significant problem in the realm of computer vision. To address this problem, we propose a fingerphoto deblurring architecture referred to as Fingerphoto Deblurring using Wavelet Style Transfer (FDWST), which aims to utilize the information transmission of Style Transfer techniques to deblur fingerphotos. Additionally, we incorporate the Discrete Wavelet Transform (DWT) for its ability to split images into different frequency bands. By combining these two techniques, we can perform Style Transfer over a wide array of wavelet frequency bands, thereby increasing the quality and variety of sharpness information transferred from sharp to blurry images. Using this technique, our model was able to drastically increase the quality of the generated fingerphotos compared to their originals, and achieve a peak matching accuracy of 0.9907 when tasked with matching a deblurred fingerphoto to its sharp counterpart, outperforming multiple other state-of-the-art deblurring and style transfer techniques.\end{abstract}

\section{Introduction}

The task of deblurring images has long been an open problem in the field of computer vision. When an image is randomly blurred, it can become very difficult to render the high-frequency components of the image. Consequently, the captured image predominantly contains lower frequencies. Fingerprint photographs, or fingerphotos, have become more popular in recent years. The ease with which a fingerphoto can be acquired is much higher than that of contact-based fingerprints. However, this ease of acquisition comes with a downside, as fingerphotos are much more subject to accidental blurring than contact-based prints. The effects of motion blur, poor camera quality, and even unstable hands can be detrimental to the quality of fingerphotos. 

Conceptually, any given image can be construed as being made up of sets of information at different frequencies, ranging from low to high. Large, monochrome, or texture-like areas are primarily made up of low-frequency information. These features have minimal variety in their pixel values and, therefore, are less prone to blurring. At the other end of the spectrum, high-frequency information refers to the edges, lines, and points of an image; sections that display a rapid transition in pixel value over a small change in location. This information is vulnerable to being lost due to blurring as the majority of pixels surrounding and making up an edge or line have drastically different values from one another. This causes high-frequency information to be dispersed into its neighboring pixels and, therefore, destroyed.

\begin{figure}
    \centering
    \includegraphics[width=0.8\linewidth]{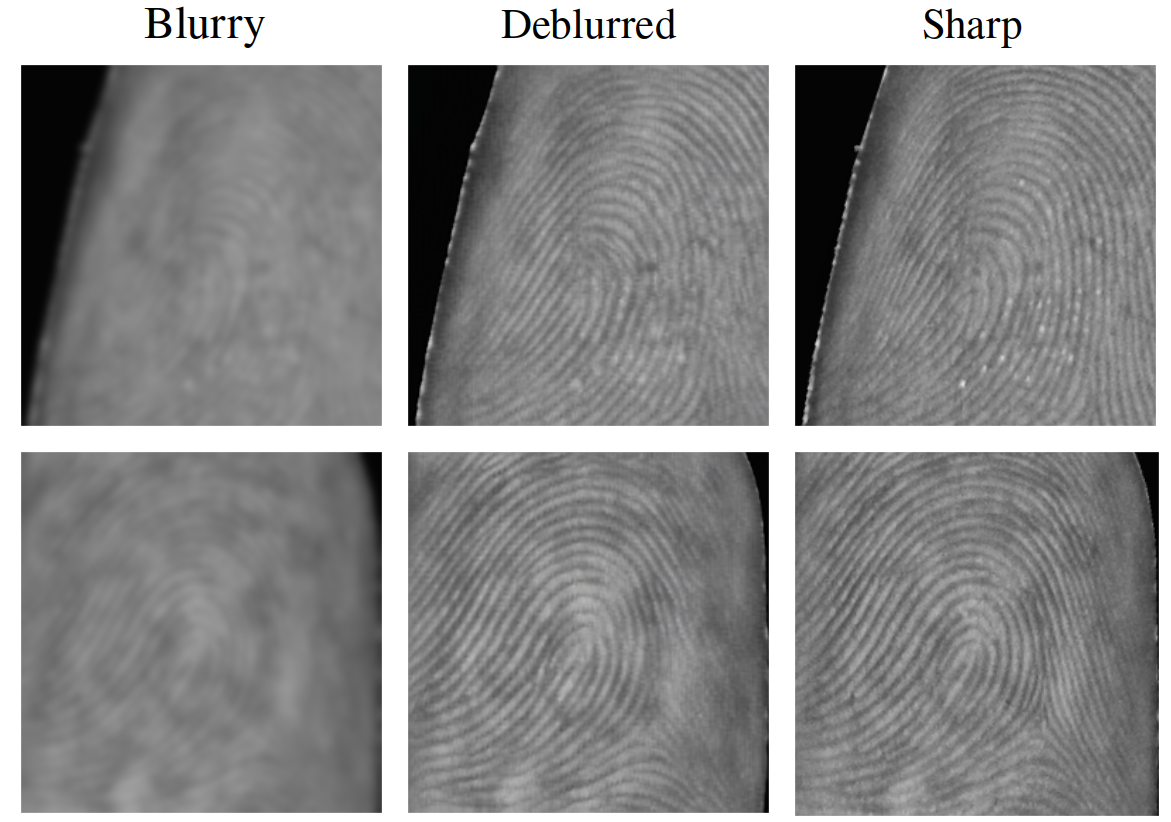}
    \caption{Sample pairs of sharp and blurry fingerphotos, with the deblurred version in the middle column.}
    \label{fig1}
\end{figure}

The inherent uncertainty involved in deblurring, mainly in the restoration of lost high-frequency information, has proven to be a daunting task for researchers. The process of recovering lost high-frequency detail is generally considered an ill-posed problem. This arises from the fact that there is no deterministic way to guarantee that a deblurred image corresponds to the unique original of an arbitrarily blurred image. The nature of fingerprints, mainly their seemingly random arrangement of ridges, loops, and bifurcations, makes them particularly formidable for classical deblurring techniques to effectively enhance. Unfortunately, these features, referred to as minutiae, comprise the bulk of the important identification information in any given fingerprint. Modern fingerprint identification systems, such as Neurotechnology's Verifinger SDK or Innovatrics' IDKit SDK \cite{veri, idk}, heavily rely on the properties of these minutiae for their fingerprint-subject verification, in addition to several other proprietary algorithms. These systems look mainly at the direction, ridge orientation, thickness, and locations of minutiae points, all of which are highly susceptible to loss through blurring. During the deblurring process, it is imperative that these identity-related features are reconstructed in a way that accurately reflects the originals, as even minor alterations to minutiae can significantly impact identification accuracy. 

In this paper, we propose a modified deep convolutional neural network termed Fingerphoto Deblurring using Wavelet Style Transfer (FDWST), which aims to combine the best aspects of both style transfer techniques and the Discrete Wavelet Transform (DWT). Our base model leverages a standard autoencoder, although with one significant modification. Unlike conventional autoencoders, we entirely discard the encoder and replace it with the DWT operation, allowing our input images to be split into distinct wavelet frequency bands. Additionally, following the DWT, we perform our Wavelet Style Transfer Module, facilitating style transfer within the wavelet domain instead of in the feature domain. By substituting the encoder for the DWT, we force our inputs to be split into distinct frequency bands, thereby enabling a more adequate transfer of sharpness information at different frequency bands. This strategy improves the overall quality of our deblurred images. 

Our proposed model goes one step further in that it can deblur any given fingerphoto with a single style image. Specifically, for every blurry fingerphoto, a single sharp fingerphoto is used. This fingerphoto undergoes binarization, i.e., every value of the pixel is set to either 0 or 1, enhancing the quality of the sharpness information. Further details on this technique are elaborated in Section 3.3. Some example results from our model are displayed in Fig. \ref{fig1}. Our primary contributions are outlined as follows:

\begin{itemize}
    \item Integration of Style Transfer and DWT in a novel approach with the goal of effectively transferring sharpness and clarity information to deblur fingerphotos.
    \item Exhibit how Wavelet Style Transfer enhances fingerphoto deblurring by facilitating the transfer of high-frequency sharpness information between images.
    \item Presentation of qualitative and quantitative evidence showcasing the superior performance of our model over other state-of-the-art deblurring and style transfer architectures.
\end{itemize}

This paper is organized into seven main sections. Section 2 discusses related work in the fields of deblurring and style transfer, as well as some other wavelet-based style transfer methods. Section 3 explains some background information needed for the model and some details on certain decisions regarding our design choices. Lastly, Sections 4, 5, 6 and 7 give a thorough overview of our implemented architecture, experiments, evaluation results, and ablation study.

\section{Related Works}
\subsection{Image Deblurring}
Image deblurring has been a fundamental task in the field of computer vision and digital graphics for many years \cite{deblur3, delur5, deblur4, deblur2, text}. The objective is straightforward; given a blurry image \(I_b\) that has been corrupted by a blur kernel \(K\) and unknown noise \(N\), restore a sharp image \(I_s\) devoid of both the effect of \(K\), as well as \(N\). There is irony in the fact that this simple concept does not have a deterministic solution, as there is inherent ambiguity involved when removing blur from an image. Despite this, models that both know (non-blind) and do not know (blind) the blur kernel have made impressive headway in recent years. Kupyn \textit{et al.} \cite{deblurgan} use a GAN architecture and simulate realistic motion blur to reverse the effects of motion blur in natural images. Chiu and Gurari \cite{blind} introduce an interesting approach, involving two GANs, one of which learns to blur and the other to deblur, and they compete with each other in such a way that fake deblurred images cannot be determined from real sharp ones. Since these deblurred images cannot be distinguished from real sharp ones, they are, by proxy, sharp themselves. Joshi \textit{et al.} \cite{joshi2023fingerphoto} incorporate multi-stage GAN architecture to deblur fingerphotos. However, the convoluted network requires a specific resolution to operate, making it challenging to use in a real-world scenario. 

\subsection{Neural Style Transfer}
Style transfer refers to the task of amalgamating two images, the content image and the style image. The objective is to combine them in such a way that the overall structure of the content image is retained, while reconstructing it in the same artistic essence as the style image. Generally, style transfer architectures are designed with the goal of mimicking an art, and not for a more practical task such as deblurring. Beginning with the landmark work of Gatys \textit{et al.} \cite{Gatys} in 2015, neural style transfer has seen an explosion of interest recently, \cite{adain3, adain4, adain2, SANet} with the growing popularity of artificially generated artwork. Gatys \textit{et al.} introduced style transfer by utilizing the Gram matrix to perform a form of style loss between two images. Huang and Belongie \cite{AdaIN} pioneered a new approach, utilizing Adaptive Instance Normalization, aligning the mean and standard deviation of the content image with the style image and thereby transferring the style. \cite{AdaAttN, SANet} expanded upon the ideas of \cite{AdaIN}, introducing attention and a form of weighted mean-variance maps, respectively. Deng \textit{et al.} \cite{stytr2} make use of transformers to generate domain-specific content and style sequences.

\subsection{Wavelet-Infused Networks}
The use of wavelets in style transfer has seen very little exploration in recent years, despite its compatibility with neural networks \cite{wavelet2, wavelet3, wavelet4}. The usefulness of the discrete wavelet transform in extracting specific frequency information from an image naturally fits into the realm of style transfer. Yoo \textit{et al.} \cite{wct2} were able to implement a form of wavelet pooling and unpooling in a photo-realistic style transfer model, improving upon existing models by reducing computation costs and decreasing the generation of unwanted image artifacts. Deng \textit{et al.} \cite{SWT} was able to use the stationary wavelet transform (SWT), instead of the DWT, to perform single-image super-resolution. 

\section{Methodology}

\begin{figure}
    \centering
    \includegraphics[width = 0.85\linewidth]{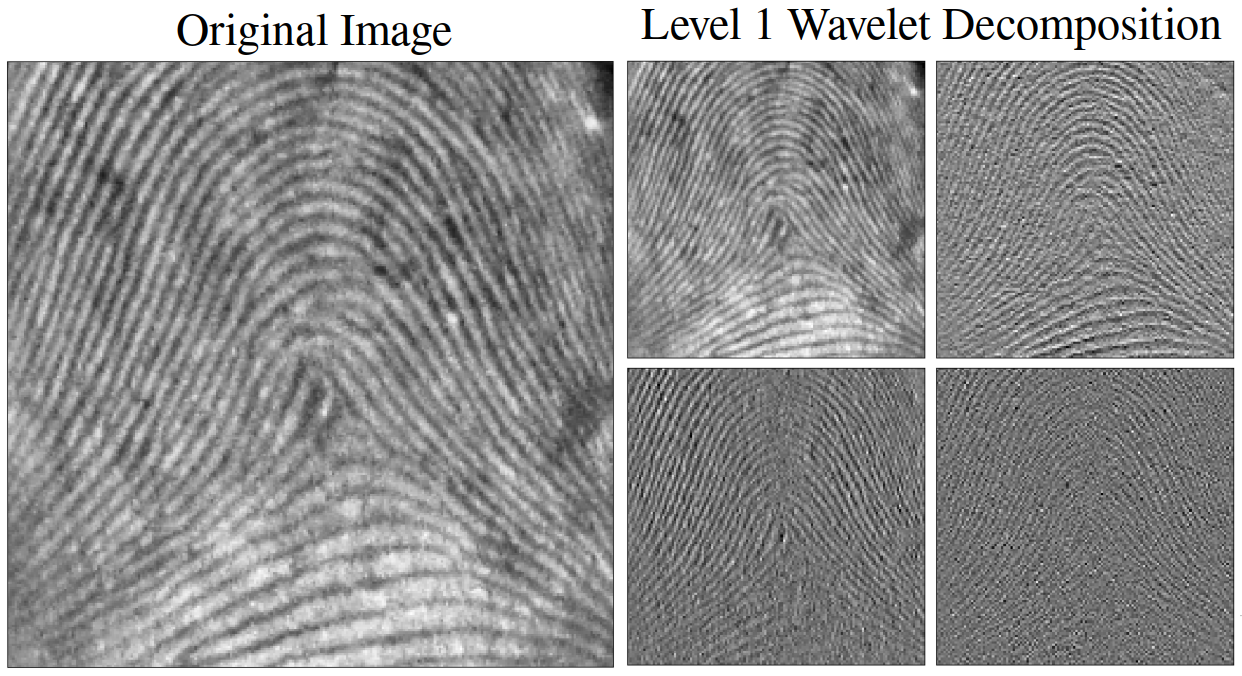}
    \caption{A sharp fingerphoto alongside its corresponding level 1 wavelet decomposition. The top left sub-band is an approximation of the original image, being a smaller smoothed version. The remaining three sub-bands contain horizontal, vertical, and diagonal information, respectively.}
    \label{fig2}
\end{figure}

\subsection{2D Discrete Wavelet Transform}\label{DWTsec}
At the core of our model lies the two-dimensional discrete wavelet transform. The DWT takes an image \(I\) and decomposes it into four filtered sub-bands, also known as wavelets, each of which is reduced to half the resolution of the original image. These four sub-bands are labeled as \(W_{LL}\), \(W_{LH}\), \(W_{HL}\), and \(W_{HH}\), each representing a unique set of information extracted from the original image. The subscript of each wavelet band refers to the order in which either low- or high-pass kernels are convolved over the original image, as seen in Eq. (\ref{eq1}). Since convolving an image with low-pass filters smooths the image, \(W_{LL}\) is referred to as the approximation wavelet, as it heavily preserves the original image's content while being slightly smoothed. The remaining three wavelets, \(W_{LH}\), \(W_{HL}\), and \(W_{HH}\), are called detail coefficients leveraging high-pass filters to enhance directional information. These sub-bands contain vertical, horizontal, and diagonal details, respectively. In order to derive the four sub-bands for each fingerphoto \(I\), we repeat four separate convolutions on \(I\), each with a distinct filter kernel. For our wavelet analysis, we used the Python PyWavelets package developed by Lee \textit{et al.} \cite{pywt}. The formulation of each kernel can be observed in Eq. (\ref{eq1}), and example wavelet sub-bands from a sharp fingerphoto can be seen on the right-hand side of Fig. \ref{fig2}. 

\begin{equation} \label{eq1}
\begin{split}
& K_{LL} = \frac{1}{2}
\begin{bmatrix}\frac{\sqrt{2}}{2}  \frac{\sqrt{2}}{2}\\
\end{bmatrix} * 
\begin{bmatrix}\frac{\sqrt{2}}{2}\\\frac{\sqrt{2}}{2}
\end{bmatrix},\;\;\;\;\;\;\;\;\\
& K_{LH} = \frac{1}{2}
\begin{bmatrix}\frac{\sqrt{2}}{2}  \frac{\sqrt{2}}{2}\\
\end{bmatrix} * 
\begin{bmatrix}-\frac{\sqrt{2}}{2}\\\frac{\sqrt{2}}{2}
\end{bmatrix}, \\
& K_{HL} = \frac{1}{2}
\begin{bmatrix}-\frac{\sqrt{2}}{2} & \frac{\sqrt{2}}{2}\\
\end{bmatrix} * 
\begin{bmatrix}\frac{\sqrt{2}}{2}\\\frac{\sqrt{2}}{2}
\end{bmatrix},\;\;\;\;\;\\
& K_{HH} = \frac{1}{2}
\begin{bmatrix}-\frac{\sqrt{2}}{2} & \frac{\sqrt{2}}{2}\\
\end{bmatrix} * 
\begin{bmatrix}-\frac{\sqrt{2}}{2}\\\frac{\sqrt{2}}{2}
\end{bmatrix}.
\end{split}
\end{equation}

This process of decomposing an image into its subsequent wavelets can be repeated to further decompose the sub-bands. Performing a Level 1 DWT on an image yields four resultant wavelets. Similarly, reapplying the process on these newly generated wavelets results in 16 sub-bands, referred to as a Level 2 DWT. This process can occur multiple times, as long as the image resolution allows it. For every level, the resolution of each wavelet halves, and the total number of wavelets increases by a factor of four. For example, an 8 x 8 image can only produce a maximum DWT of level 3, where each wavelet is a single pixel, discarding meaningful information such as mean and variance. Therefore, we must make a decision on which level wavelet decomposition to utilize in our network. Details about this choice are explained in Section \ref{IDWTsec}.

\subsection{Wavelet Style Transfer}\label{}\label{WSTsec}

The detail coefficients \(W_{LH}\), \(W_{HL}\), and \(W_{HH}\) from the DWT contain specific directional high-frequency information. Our proposed Wavelet Style Transfer (WST) takes advantage of this property to increase sharpness and deblur fingerphotos. First, we generate a set of level 3 DWT sub-bands from a blurry fingerphoto, \(I_b\) and a sharp one, \(I_s\): 

\begin{equation} \label{eq2}
\{W_x^{(1)}.W_x^{(2)}, \ldots , W_x^{(64)}\} = DWT(I_x).
\end{equation}

It should be noted that these two fingerphotos do not necessarily need to share the same ID; i.e, belonging to the same subject. This step results in 128 wavelets total, where 64 are from \(I_b\) and 64 are from \(I_s\). For each pair of wavelets \((W_b^{(i)}, W_s^{(i)})\), where \(i\) is the index of each wavelet in each set, we normalize each blurry wavelet \(W_b^{(i)}\) with respect to its mean \(\mu(W_b^{(i)})\) and standard deviation (STD) \(\sigma(W_b^{(i)})\) to acquire the normalized wavelet \(W_b^{(i)^\prime}\). We then multiply \(W_b^{(i)^\prime}\) by \(\sigma(W_s^{(i)})\) and add \(\mu(W_s^{(i)})\), transferring the style and high-frequency information present in \(I_s\) to \(I_b\), and creating a style-transferred set of wavelets \(W_t^{(i)}\): 

\begin{equation} \label{eq3}
W_t^{(i)} = WST(W_b^{(i)}, W_s^{(i)}),\; \text{for} \;i=1 \ldots 64.
\end{equation}

\begin{figure}
    \centering
    \includegraphics[width=0.85\linewidth]{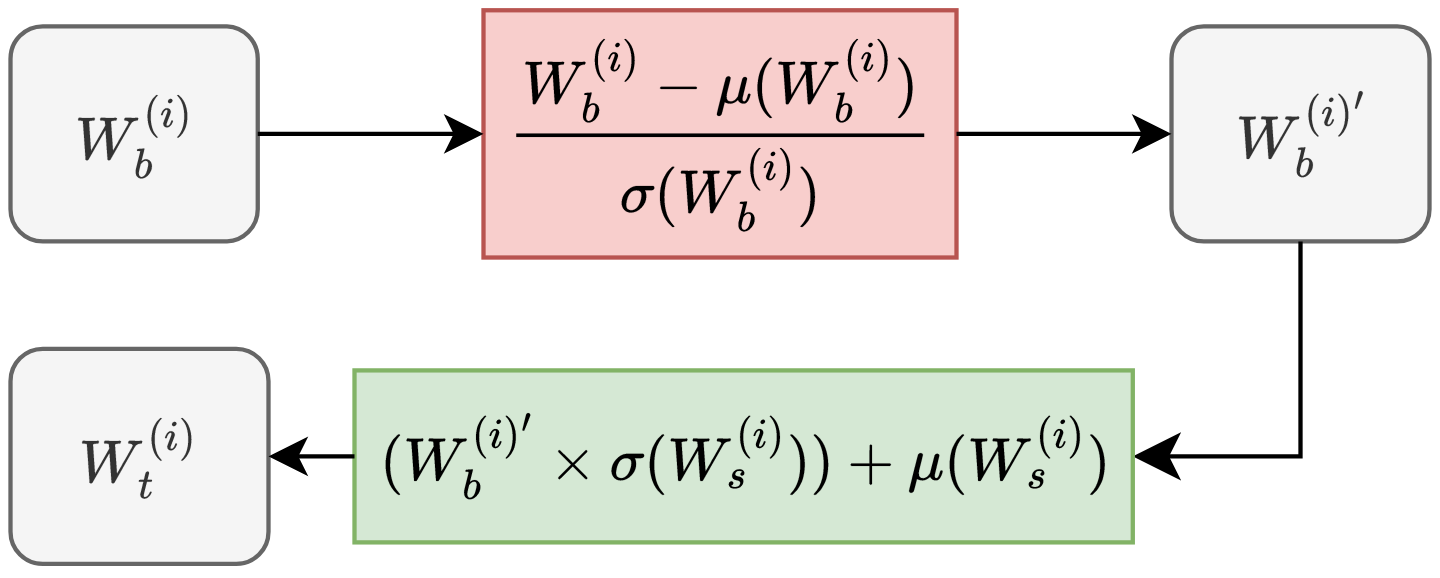}
    \caption{Our proposed Wavelet Style Transfer (WST) module, which enhances a blurry set of wavelets \(W_b^{(i)}\) with a sharp set \(W_s^{(i)}\), and produces a style transferred set \(W_t^{(i)}\).}
    \label{wst}
\end{figure}

The proposed WST grants key advantages when specifically deblurring fingerphotos. From our experiments discussed in Section \ref{IDWTsec}, it is evident that there is a significant contrast between the variance of the DWT sub-bands of a blurry and a sharp fingerphoto. Higher variance in the sub-bands is a consequence of varying gradients due to the ridges and valleys in the fingerphotos. Therefore, we can confidently hypothesize that our WST is transferring sharpness information, in the form of mean and STD, from a sharp fingerphoto to a blurry one. The proposed module can be seen in detail in Fig. \ref{wst}. 

\subsection{DWT Levels}\label{IDWTsec}

As stated in Sections \ref{DWTsec} and \ref{WSTsec}, we employ a Level 3 DWT for our transfer process. This is because, depending on the selected DWT level, our deblurred images may exhibit one of two undesirable qualities upon reconstruction from the Inverse Discrete Wavelet Transform (IDWT). A Level 2 or lower DWT fails to sufficiently sharpen the output images, giving results that remain excessively blurry. Conversely, with higher levels of DWT, our deblurring results contained several artifacts. For example, decomposing the image into smaller sub-bands allows the sharp image to leak its identity information into the blurry one. Considering that we are using fingerphotos belonging to two different individuals, this causes the two separate identities to mix into a single image, thereby compromising the identity of the original image and rendering our results unusable. 

Considering that all our training fingerphotos are scaled to size $256 \times 256$, and each image contains \(256^2 = 65,536\) pixels, the maximum feasible DWT level is level 8. Since \((\frac{1}{4})^8=\frac{1}{65,536}\), and the wavelet resolution halves with each level, a level 8 DWT yields \(4^8 = 65,536\) wavelets at a size of a single pixel each. Consequently, the mean of each wavelet is insignificant since each wavelet is a single pixel, and therefore, when we subtract \(W_b^{(i)}\) and add \(W_s^{(i)}\) during WST, every blurry wavelet is completely replaced by its corresponding sharp wavelet. This property can be observed at the bottom right of Fig. \ref{levels}. This ID-replacement phenomenon intensifies with the DWT level from 1 to 8, but retained blurriness increases from level 8 to 1. Additionally, each level of DWT augments the total number of feature maps fed into the decoder by a factor of 4, as our wavelets are fed directly into the network. Therefore, we utilize level 3 DWT for our WST. This choice exhibited a reasonable amount of enhancement, while mitigating artifacts and maintaining the identity of the original image. Moreover, it maintains a realistic total number of wavelets. Examples of higher- and lower-level IDWT results can be observed in Fig. \ref{levels}.

\begin{figure}
    \centering
    \includegraphics[width=0.85\linewidth]{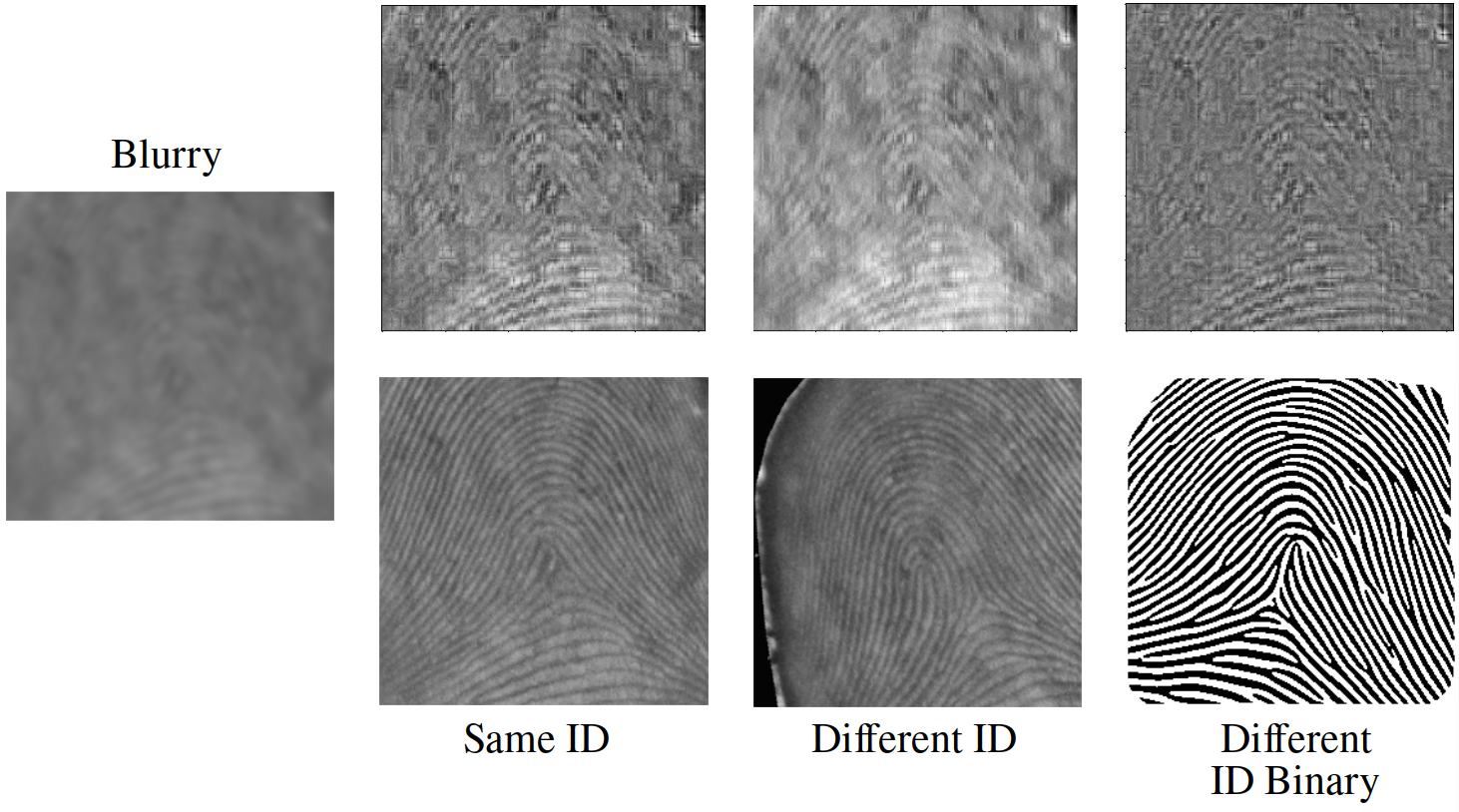}
    \caption{Sample IDWT results from a Level 3 DWT. The Blurry photo on the left is fixed, and we perform WST with three different sharp images before inverting. The first sharp image is the same ID fingerphoto as the blurry one, the second is a different ID, and the third is a binarized different ID.}
    \label{inverses}
\end{figure}

\begin{figure}
    \centering
    \includegraphics[width=0.85\linewidth]{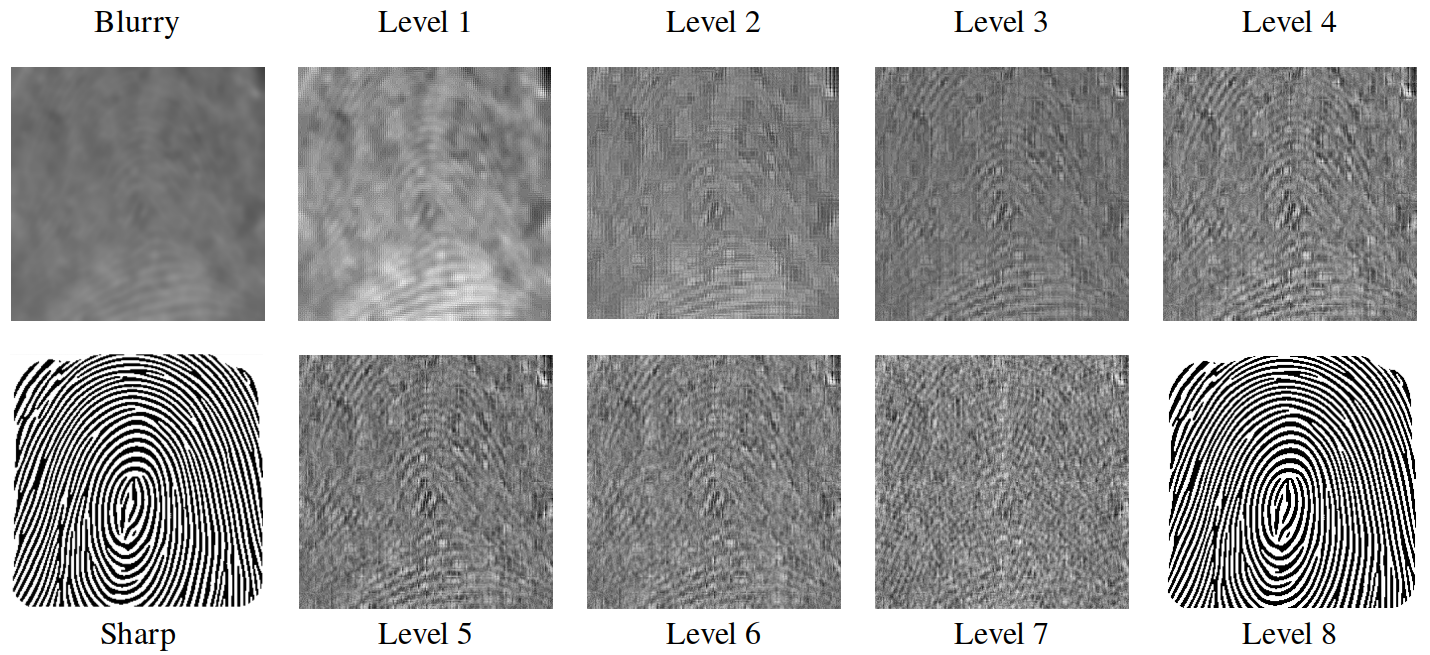}
    \caption{Examples of IDWT reconstructions from different level DWTs. The lower the level, the blurrier the reconstruction. With higher levels, artifacts become prevalent and the identities mix. Finally, at Level 8, the sharp identity fully replaces the original blurry one.}
    \label{levels}
\end{figure}

\begin{figure}
    \centering
    \includegraphics[width=0.4\linewidth]{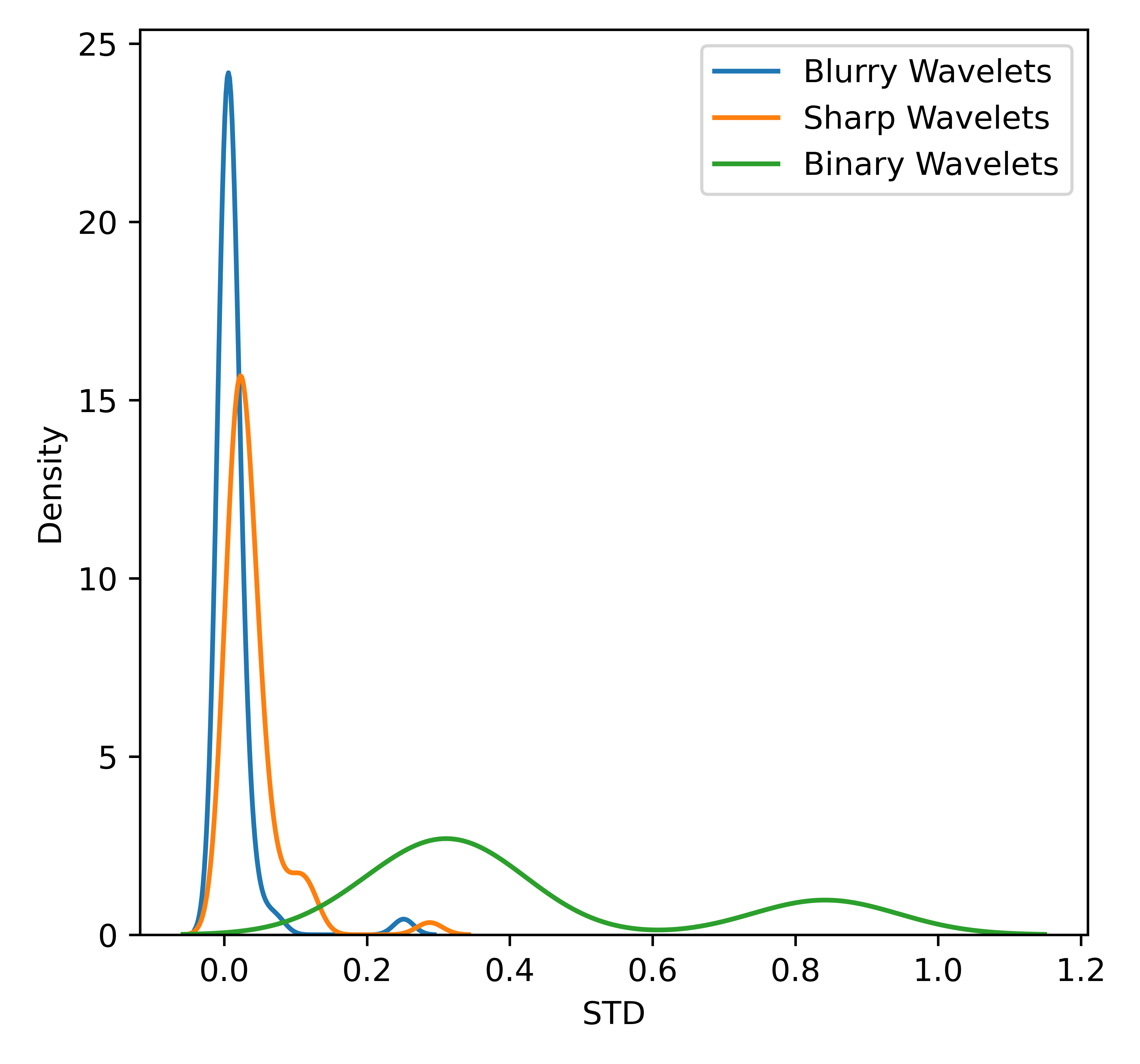}
    \caption{Density plot displaying the standard deviations of each of the sub-bands in \(W_x^{(i)}\) for a sample fingerphoto, where \(x\) is either blurry \((b)\), sharp \((s)\), or binary \((bin)\). The binarized fingerphoto clearly displays higher variance among its wavelets, and therefore contains higher-frequency information than the grayscale sharp or blurry version.}
    \label{density}
\end{figure}

\subsection{Binary vs. Sharp Fingerphoto}

\begin{figure*}
\begin{center}
\includegraphics[width=0.9\linewidth]{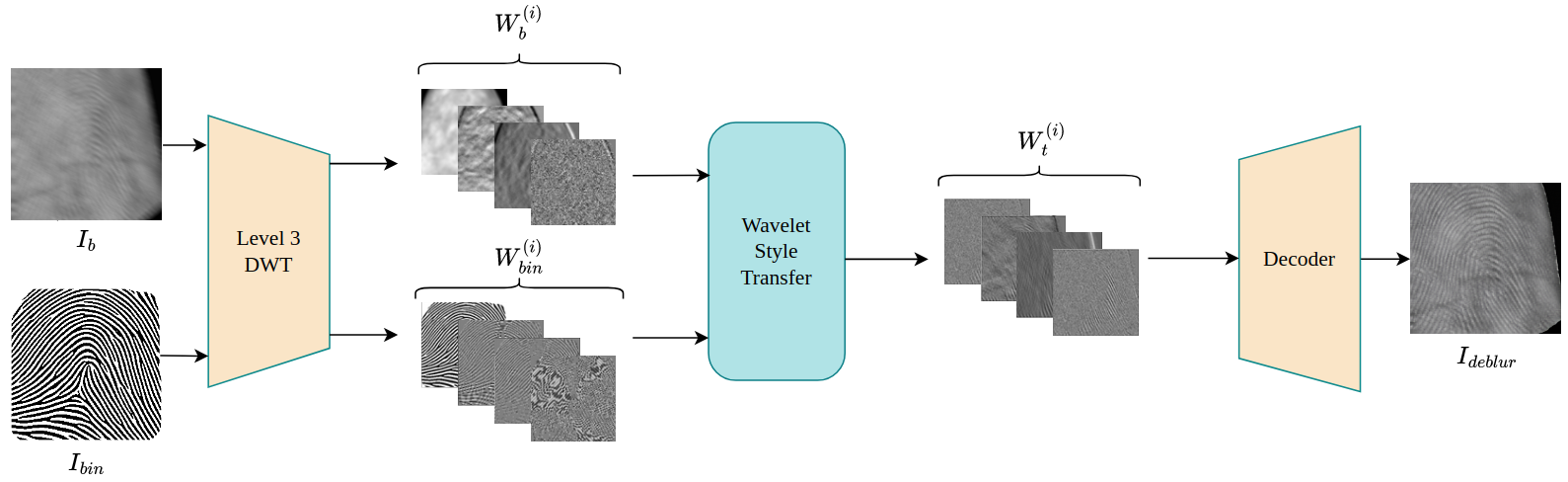}
\end{center}
   \caption{Our proposed deblurring model. Two input fingerphotos \(I_b\) and \(I_{bin}\) are first passed into the DWT, producing two sets of wavelets \(W_b^{(i)}\) and \(W_{bin}^{(i)}\). The resulting wavelets are merged, transferring the style from \(I_{bin}\) onto \(I_b\). We then treat this combined set of wavelets \(W_{t}^{(i)}\) as feature maps for our decoder, which combines them back into a deblurred fingerphoto \(I_{deblur}\).}
\label{model}
\end{figure*}

The selection of the sharp fingerphoto for WST is crucial because it directly impacts the quality of the deblurring process. We observed that DWT sub-bands with higher variance indicate a higher level of sharpness in the original fingerphoto. Hence, we advocate using an image with the highest possible variance, or sharpness, for deblurring. To this aim, we initially examined the STD of \(W_b^{(i)}\) and \(W_s^{(i)}\), displayed in Fig. \ref{density}. We can observe that the wavelets \(W_s^{(i)}\) contain a slightly higher variance than those of \(W_b^{(i)}\). This disparity is expected, and attributed to the lower frequency information present in blurry images. Based on this observation, as well as results using DWT, we devised a strategy to binarize an arbitrary sharp fingerphoto \(I_s\) into \(I_{bin}\). This binarized fingerphoto was taken from the VeriFinger SDK. 

The STD of this new \(I_{bin}\) is significantly higher than that of both \(I_b\) and \(I_s\), indicating that theoretically it has distinctive deblurring information present. The resulting IDWT-reconstructed fingerphotos, now subjected to style-transfer with the new binarized fingerphoto, exhibit a degree of embedded sharpness but also manifest jagged artifacts. The results showing IDWT reconstructions using different style-transfer fingerphotos are provided in Fig. \ref{inverses}. Since the WST is adept at performing sharpness-enhancing style transfer without necessitating an encoder, we directly pass the new style-transferred wavelet set \(W_{t}^{(i)}\) to our decoder for subsequent reconstruction and refinement.

\section{Architecture}

\subsection{Model}
Our proposed model is comprised of three main modules. Firstly, we employ a Level 3 DWT that serves as the encoder for our network. Secondly, we implement our WST module to execute style transfer and prepare the wavelets for the network. Finally, we include a decoder as our last segment. The decoder consists of a total of 9 layers. Layer 1 accepts 64 input channels, corresponding to our 64 generated wavelets, and produces 512 output channels. Subsequently, layers 2, 5, and 7 halve the total number of channels, starting from 512, and reduce them to 256, 128, and eventually back to 64 channels. All convolution kernels utilized are of size 3x3. The Rectified Linear Unit (ReLU) activation function and a 2D reflection pad are included between each layer. Additionally, layers 1, 5, and 7 incorporate an upsample operation to increase spatial size back to the original.

\subsection{Loss Functions}
Our network combines three loss functions for its training. They are content, style, and identity loss:

\begin{equation} \label{losstotal}
\mathcal{L}_{total} = \lambda_c\mathcal{L}_c + \lambda_s\mathcal{L}_s + \lambda_i\mathcal{L}_i,
\end{equation}
where, \(\mathcal{L}_c\) is content loss, \(\mathcal{L}_s\) is style loss, and \(\mathcal{L}_i\) is identity loss. \(\lambda_c\), \(\lambda_s\) and \(\lambda_i\) are hyperparameters used to weigh each loss with respect to the others. For our final model, our loss weights were set to \(\lambda_c = 100\), \(\lambda_s = 10\), and \(\lambda_i = 25\).

\subsubsection{Content Loss}
As it is beneficial in image generation tasks, we utilize L1 loss between the original sharp fingerphoto \(I_s\) and its deblurred version \(I_{deblur}\), over a batch of size \(n\), in order to retain as much identity information as we can upon deblurring. This helps remove arbitrarily generated minutiae that do not reflect \(I_s\) and preserve the original content image as much as possible while simultaneously sharpening edges, as shown here:

\begin{equation} \label{losscontent}
\mathcal{L}_c = \frac{1}{n} \sum_1^n |I_{deblur}^{(n)} - I_s^{(n)}|. 
\end{equation}

\subsubsection{Style Loss}
We incorporate style loss using the generated feature maps of a pretrained VGG19 encoder. To better retain both local and global style features from our deblurred images, we use all 5 activation layers of the VGG encoder for our feature map generation. Inspired by \cite{AdaIN}, we then take the feature representations of both \(I_s\) and \(I_{deblur}\), and calculate the difference between their mean and STD:

\begin{equation} \label{lossstyle}
\begin{split}
&\mathcal{L}_{s}=\sum_{\phi = 1}^5
\left|\left|\mu(V_{\phi}(I_{deblur}))-\mu(V_{\phi}(I_s))\right|\right|_2
\\
&\;\;\;\;+\; \sum_{\phi = 1}^5\left|\left|\sigma(V_{\phi}(I_{deblur}))-\sigma(V_{\phi}(I_s))\right|\right|_2.
\end{split}
\end{equation}

Here, \(V_{\phi}(.)\) represents the feature maps generated by the corresponding \(\phi^{th}\) activation layer of a pretrained VGG19 encoder.

\subsubsection{Identity Loss}
For our Identity Loss, we incorporated a separate pre-trained ResNet-18 verifier to preserve ID information throughout the deblurring process. This verifier is trained to match fingerphotos in the embedded domain and extracts multi-level features of a given fingerphoto. We take the Mean Squared Error (MSE) loss between the embeddings, and then between the features of \(I_s\) and \(I_{deblur}\):

\begin{equation} \label{lossidentity}
\begin{split}
\mathcal{L}_{s}=&
\left|\left|E(I_{deblur})-E(I_s)\right|\right|_2
\\
&+ \sum_{i = 1}^3
\left|\left|F_{i}(I_{deblur})-F_{i}(I_s)\right|\right|_2,
\end{split}
\end{equation}
where, \(E\) represents the embeddings from the last layer of the verifier, and \(F_i\) represents the generated features at the \(i^{th}\) intermediate layer within the verifier. This helps ensure that the features of \(I_{deblur}\), or the minutiae, closely match those of \(I_s\), subsequently maintaining identity through transfer.

\section{Experiments}
\subsection{Training}
For our training, we trained the model for a total of 240 epochs. We use the Adam optimizer, with \(\beta_1 = 0.9\), \(\beta_2 = 0.999\), and an initial learning rate of 0.0001. Our model was trained on three NVIDIA GeForce GTX 1080 Ti.

For our training dataset, we used the WVU Multimodal Fingerphoto Dataset \cite{biocop}. This dataset consists of 3,542 fingerphotos, all of which have a resolution $256 \times 256$. We blurred each fingerphoto during training, using one of three equally likely blurring methods: Gaussian, Motion, or Average. It should be noted that the kernel size for all three blur types is given as \(k = 6\sigma - 1\) where \(\sigma\) is an integer from range \([3, 6]\). We chose Gaussian Blur's \(\sigma\) to be in range \([3, 6]\), and Average Blur to use a kernel size of \(\frac{k}{2}\), as a kernel size of \(k\) produced images that were unrealistically blurry to contain any relevant identifying information. 

\subsection{Evaluation}
Since our method integrates both style transfer and deblurring, we compare our method with other style transfer and deblurring architectures. To measure our performance, we fed our deblurred results to both Neorutechnology’s Verifinger and Innovatrics’ IDKit to obtain evaluations from multiple matchers. Our primary evaluation metrics comprise Receiver Operating Characteristic (ROC) curves, and are Area Under the Curve (AUC), and Equal Error Rate (EER). ROC assesses the model's ability to classify deblurred fingerphotos into their sharp counterparts (genuine pairs) as well as its capacity to distinguish imposter pairs, where a sharp fingerphoto belongs to a different identity. The AUC can be interpreted as a measure of the accuracy of the model. Additionally, we use the EER metric, which indicates the point at which the model experiences equal occurrences of False Positives and False Negatives. In addition to ROC curves, we also used NFIQ2 V2.2, from NIST \cite{nfiq} to evaluate the quality of our generated fingerphotos compared to their originals. By combining both quality and verifier evaluations, we can ensure that our deblurred fingerphotos retain their identity and increase sharpness. 


\begin{figure*}
     \centering
     \begin{subfigure}[b]{0.3\textwidth}
         \centering
         \includegraphics[width=\textwidth]{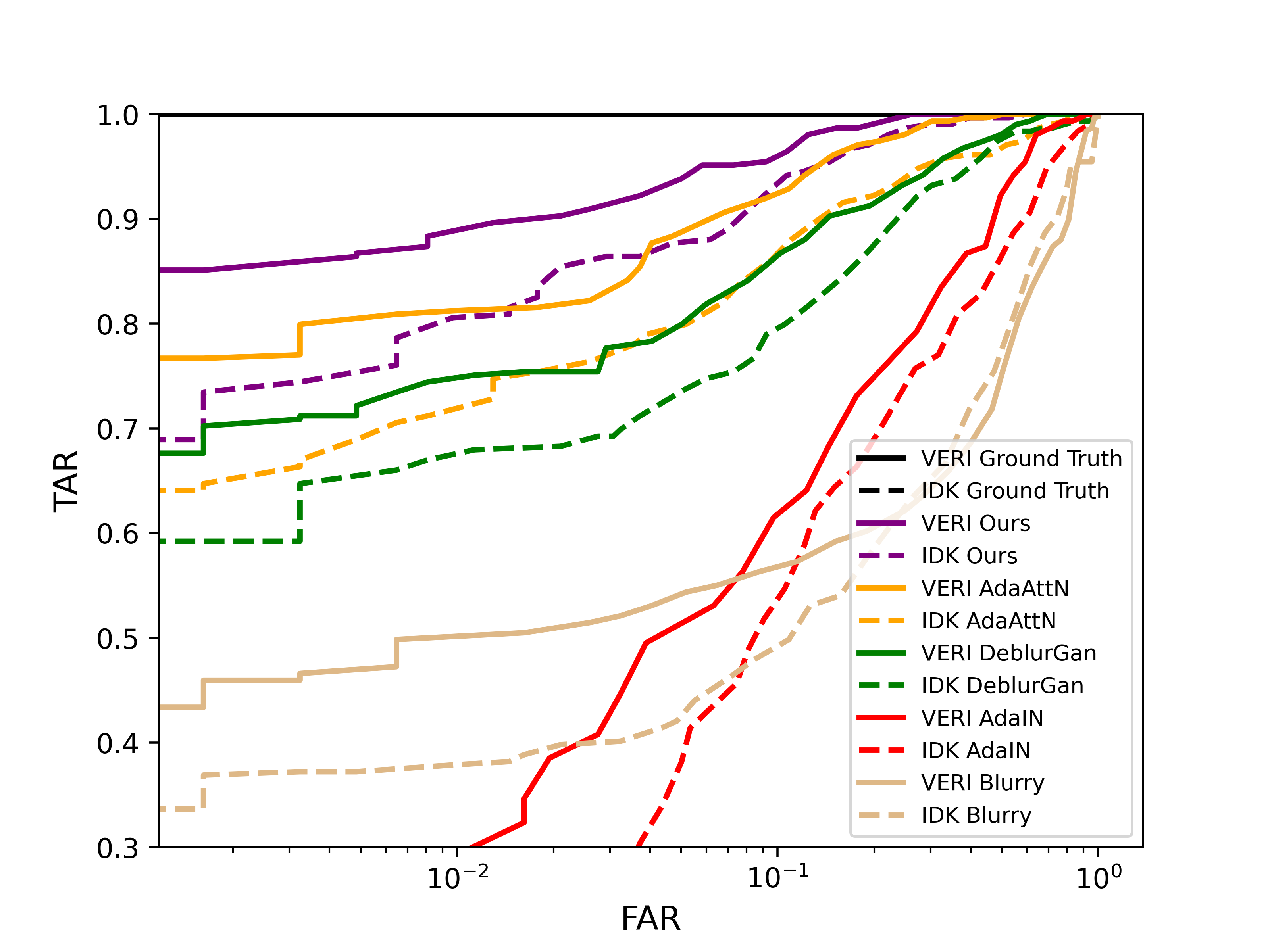}
         \caption{Multimodal}\label{a}
     \end{subfigure}
     \hfill
     \begin{subfigure}[b]{0.3\textwidth}
         \centering
         \includegraphics[width=\textwidth]{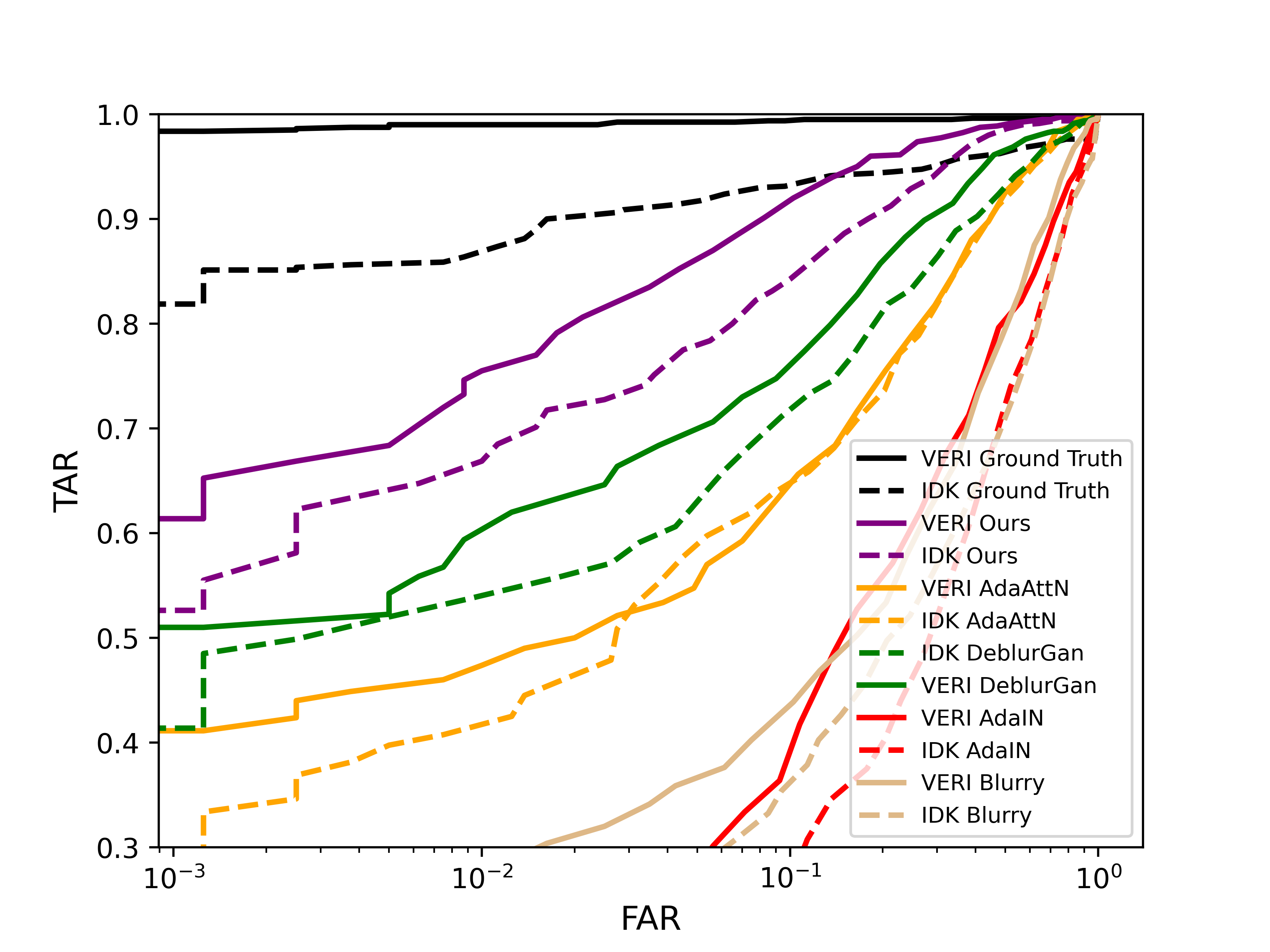}
         \caption{IIT-B}\label{b}
     \end{subfigure}
     \hfill
     \begin{subfigure}[b]{0.3\textwidth}
         \centering
         \includegraphics[width=\textwidth]{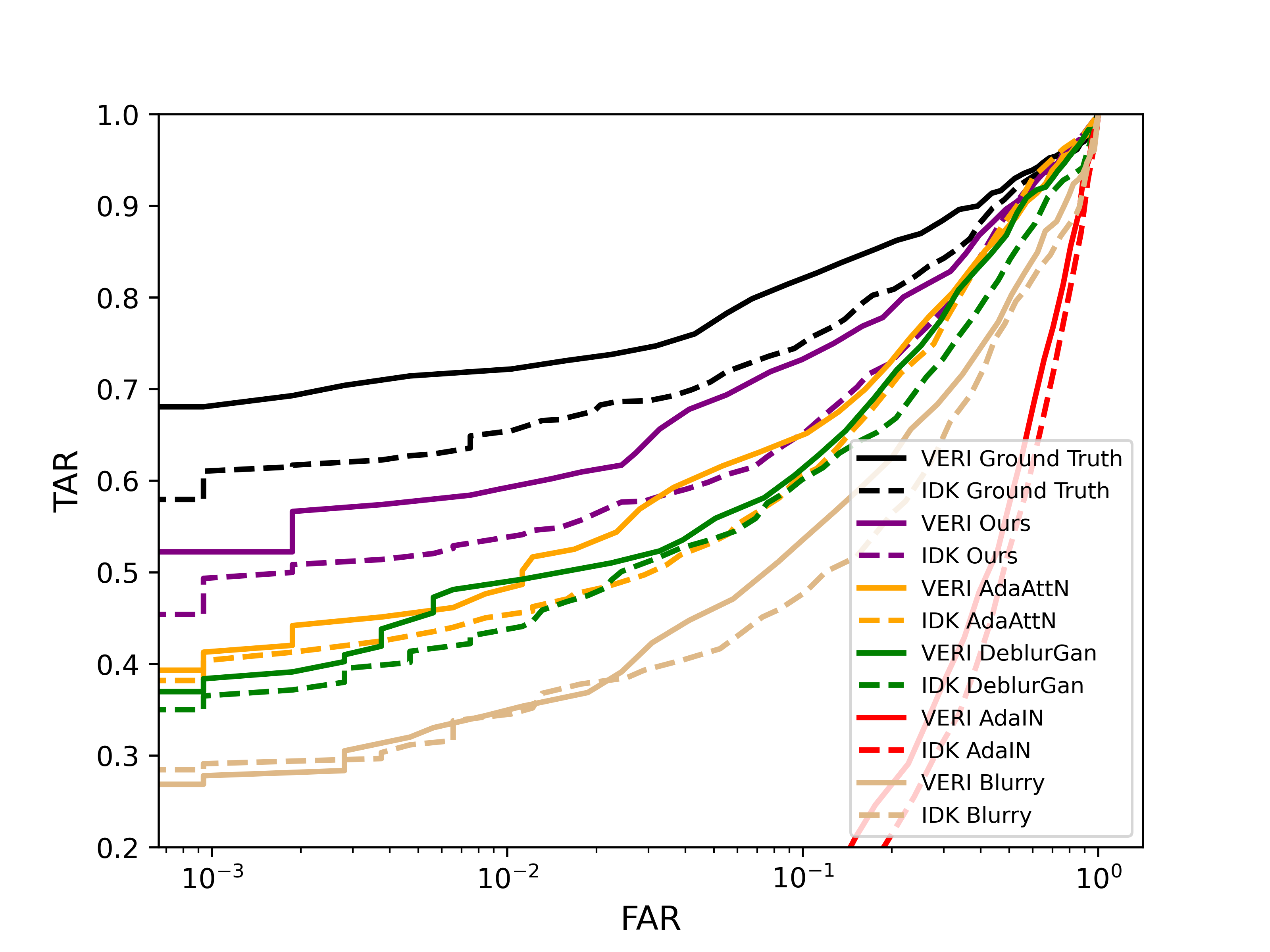}
         \caption{Ridgebase}\label{c}
     \end{subfigure}
        \caption{ROC curves on three Datasets: (a) WVU Multimodal, (b) IIT-Bombay, (c) Ridgebase. Solid lines correspond to VeriFinger results, and dashed lines represent IDKit results. Best seen in color. Zoom in for better view.}
        \label{results}
\end{figure*}

\begin{table*}
\begin{center}
\begin{tabular}{|l|c|c|c|c|c|c|}
\hline
& \multicolumn{3}{|c|}{IDKit} & \multicolumn{3}{|c|}{VeriFinger}\\
\hline
& Multimodal & IIT-B & Ridgebase & Multimodal & IIT-B & Ridgebase \\
\hline
Model & AUC/EER & AUC/EER & AUC/EER & AUC/EER & AUC/EER & AUC/EER \\
\hline\hline 
Ground Truth & $1.0000/0.0000$ & $0.9583/0.0725$ & $0.8835/0.1931$ & $1.0000/0.0000$ & $0.9958/0.0100$ & $0.9075/0.1548$ \\
Blurry       & $0.7514/0.3334$ & $0.6821/0.3746$ & $0.7336/0.3285$ & $0.7565/0.3407$ & $0.7477/0.3379$ & $0.7647/0.3057$ \\
AdaIN        & $0.8106/0.2549$ & $0.6559/0.3868$ & $0.5095/0.4876$ & $0.8575/0.2312$ & $0.7347/0.3227$ & $0.5556/0.4654$ \\
AdaAttN      & $0.9506/0.1154$ & $0.8622/0.2271$ & $0.8366/0.2593$ & $0.9812/0.0844$ & $0.8669/0.2271$ & $0.8442/0.2389$ \\
DeblurGAN    & $0.9315/0.1571$ & $0.8888/0.1939$ & $0.8039/0.2779$ & $0.9548/0.1213$ & $0.9185/0.1698$ & $0.8334/0.2521$ \\
FDWST & $\textbf{0.9795/0.0852}$ & $\textbf{0.9483/0.1297}$ & $\textbf{0.8720/0.2074}$ & $\textbf{0.9907/0.0545}$ & $\textbf{0.9703/0.0908}$ & $\textbf{0.8720/0.2074}$ \\

\hline
\end{tabular}
\end{center}
\caption{Quantitative results comparing our FDWST against other models, over all three datasets WVU Multimodal, IIT-Bombay, and Ridgebase, and from two verifiers, VeriFinger and IDKit.}
\end{table*}

\section{Results}
We evaluated our model on multiple different test sets with variations in blurring. Three test sets were used, the Multimodal Dataset \cite{biocop}, from West Virginia University, the Touch and Touchless Fingerprint Dataset \cite{iitb} from the Indian Institute of Technology, Bombay, and the RidgeBase Benchmark Dataset \cite{ridgebase}, from the University of Buffalo. These test sets contained 309, 800, and 2,229 images, respectively. For testing, we created genuine pairs using different instances of the same finger, and for impostor pairs, we randomly selected a sample from a different subject.

\begin{figure}
    \centering
    \includegraphics[width=1\columnwidth]{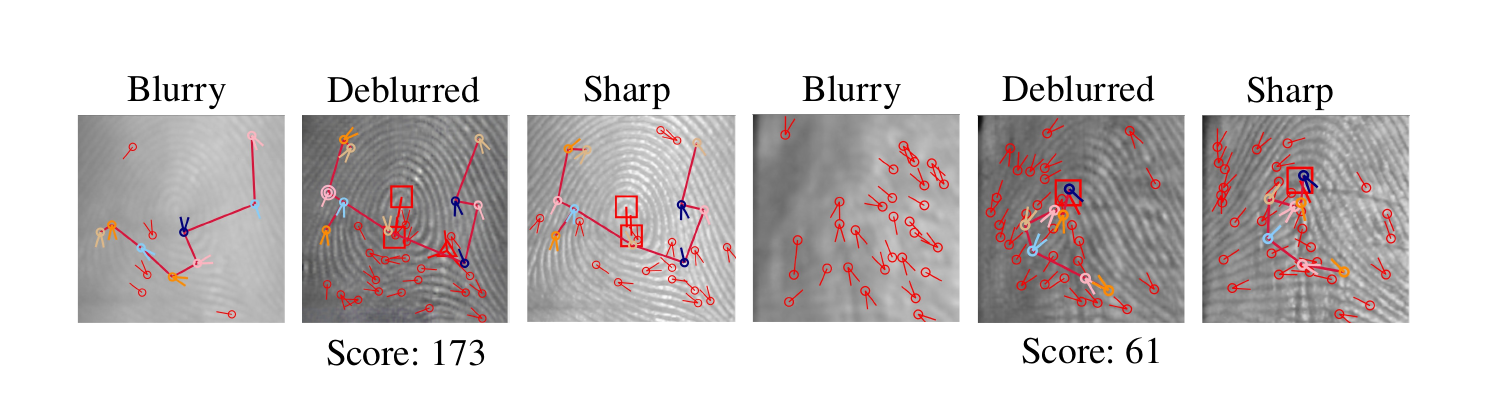}
    \caption{Naturally blurred fingerphotos, with their corresponding minutiae maps from VeriFinger SDK. The scores at the bottom of the figure correspond to deblurred and sharp pair. VeriFinger failed to generate matching score for blurry and deblurred pair.}
    \label{real}
\end{figure}

\begin{figure}
    \centering
    \includegraphics[width=1\linewidth]{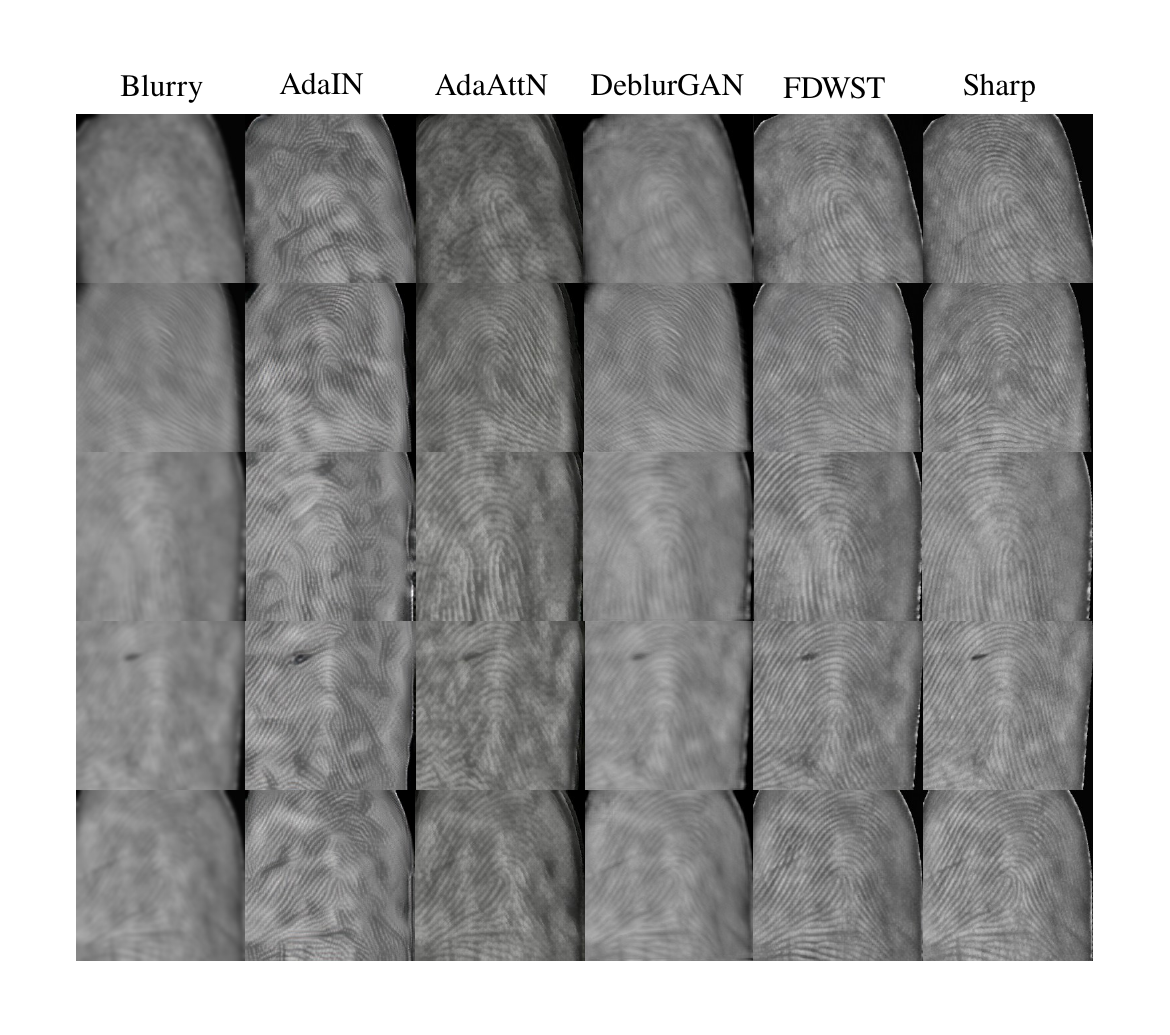}
    \caption{Sample fingerphotos from each model, as well as the original sharp and blurry versions.}
    \label{fig:grid_results}
\end{figure}

As depicted in Fig. 8 and Table. 1, our FDWST outperforms other style transfer and deblurring methods when deblurring fingerphotos. Due to the absence of proper blurry-sharp pairs for testing, we initially blurred each dataset and then passed these images through each model to obtain our results. The results in Figs. \ref{results} and \ref{fig:grid_results}, which indicate that the FDWST achieves the highest matching accuracy (AUC), coupled with a significant increase in image quality upon deblurring as shown in Table. 2, reaffirms that our model deblurs fingerphotos in such a way that retains their original identity while also generating realistic minutiae. In addition, we also evaluated our model on available naturally blurred samples. These results are depicted in Fig. \ref{real}. It can be observed that the original blurry images contain numerous incorrect minutiae, or none at all, whereas our model is able to accurately reconstruct relevant minutiae.

\begin{table}
\begin{center}
\begin{tabular}{|l|c|c|}
\hline
Dataset & Before & After \\
\hline\hline
WVU Multimodal & $0.5954$ & $23.2168$ \\
IIT-Bombay & $12.7204$ & $39.4500$ \\
RidgeBase & $15.2445$ & $25.6419$ \\
\hline
\end{tabular}\label{quality_table}
\end{center}
\caption{Quality Scores of different datasets before and after being passed through our proposed model. Each dataset shows a substantial increase in quality after the deblurring process.}
\end{table}

\section{Ablation Study}\label{ablation-sec}
For our ablation study, we examined the impact of removing different loss terms from the network, as well as comparing performance by eliminating the decoder and employing the IDWT operation for reconstruction.

\subsection{Stand-alone IDWT}\label{stand-alone-sec}
Initially we experimented with the Inverse Discrete Wavelet Transform (IDWT). Since DWT is a linear transformation, any set of wavelets \(W_{I}^{(i)}\) from image \(I\) can be used to perfectly reconstruct the original image using IDWT. First, we performed the WST operation on \((W_b^{(i)}, W_s^{(i)})\) to achieve a style-transferred set of wavelets \(W_t^{(i)}\), and then fed these new wavelets to the IDWT for reconstruction. We noticed that this linear process helps to deblur images. Furthermore, any sharp \(I_s\) could be used to enhance blurred \(I_b\) indicating that sharpness information is being transferred between the two images. To examine the effectiveness of our WST operation, we fed resultant images from the IDWT operation directly to VeriFinger, and attempted to match them to their sharp counterparts. We first investigated the performance using a grayscale fingerphoto for the style transfer process, and then swapped it out for a binarized image. As seen in Fig. \ref{ablation_pic}, the WST process followed by IDWT, notably without a decoder network, performs well. 

As illustrated in Table \ref{ablation-table}, utilizing a binarized fingerphoto as opposed to a sharp one yields a significant improvement in performance. Our binary fingerphoto led us to a \(~2.3\%\) increase in AUC, as well as a \(50\%\) drop in EER, compared to reconstructing with a sharp grayscale fingerphoto. This outcome influenced our decision to use a binarized fingerphoto to perform style transfer in our final proposed model. The IDWT operation surpasses our models with certain loss terms removed, exhibiting better performance than both our content loss and style loss removed models. With just Identity Loss removed, the binary IDWT operation performed nearly identically to our model, with changes of only 0.0016 and 0.0008 in AUC and EER, respectively. We believe that these results support the claim that our WST operation is transferring high-frequency and sharpness information from a binarized fingerphoto to a blurry one.

\subsection{Removed Loss Terms}
The removal of any loss term from our network reduced performance, with some having more significant impacts than others. In particular, eliminating identity loss did not greatly affect performance, which was anticipated given that identity loss plays a smaller role than either content or style loss. The purpose of identity constraint in our network is to catch and correct minor minutiae discrepancies between deblurred and sharp fingerphotos, so its removal does not affect the network as broadly as the other loss terms. 

\begin{figure}
    \centering
    \includegraphics[width=0.9\linewidth]{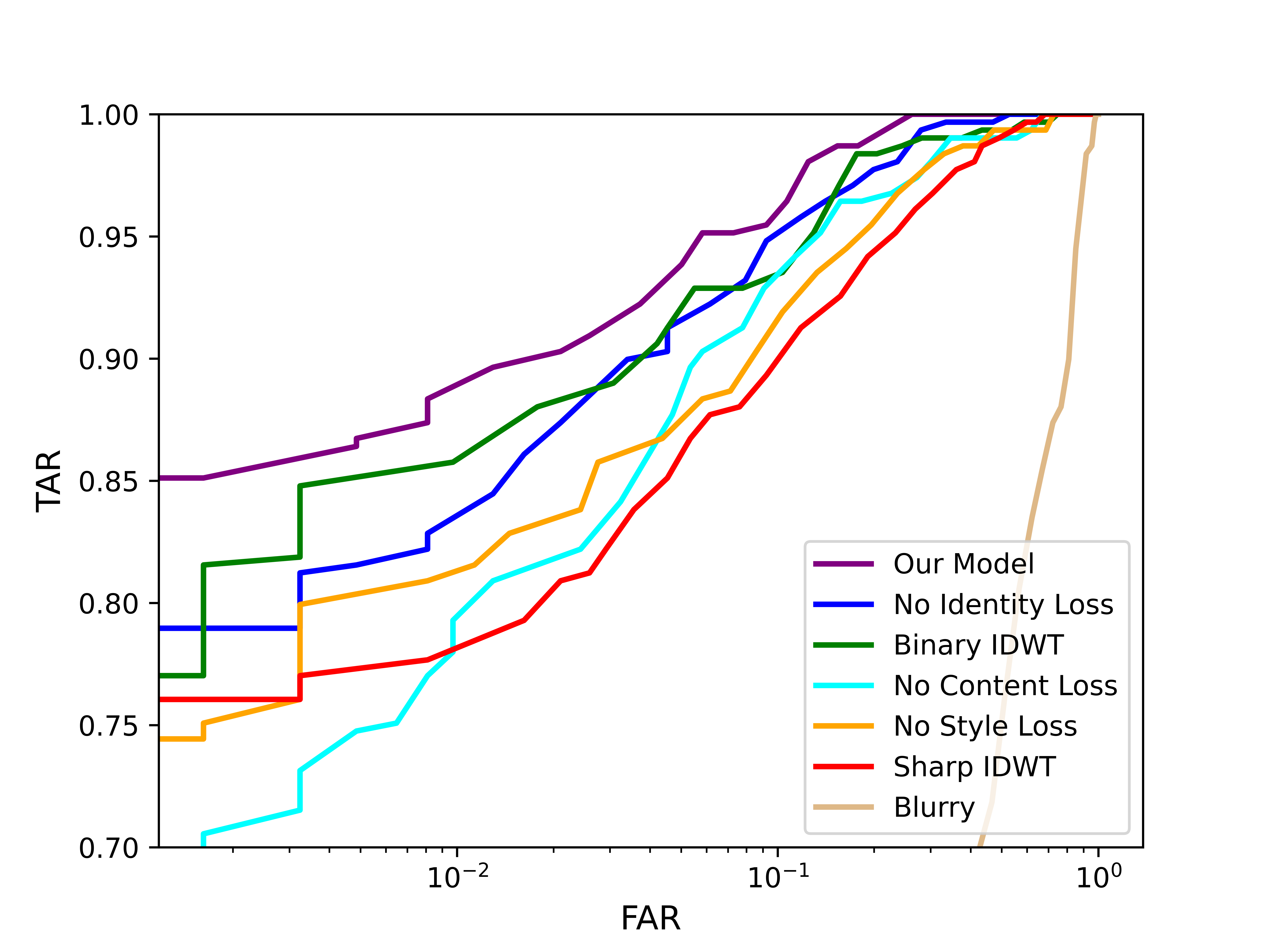}
    \caption{ROC curve results from different versions of the model obtained using VeriFinger SDK.}
    \label{ablation_pic}
\end{figure}

However, without content loss, our network produced visually appealing images that lost a significant portion of their identity. This underscores the critical role of content loss in preserving the identity of the generated fingerphotos. Furthermore, removing style loss from the network resulted in fingerphotos that did not achieve a desirable level of deblurring in terms of overall structure and sharpness. This observation suggests that style loss, since it is calculated on the basis of mean and STD, measures the level of sharpness transferred during WST. A comprehensive comparison of all loss-removed models can be seen in Fig. \ref{ablation_pic} and Table \ref{ablation-table}.

\begin{table}
\begin{center}
\begin{tabular}{|l|c|c|}
\hline
Model & AUC & EER \\
\hline\hline
    
    Our Deblurring Model & $\textbf{0.9907}$ & $\textbf{0.0545}$ \\
    Model Without Identity Loss & $0.9847$ & $0.0719$ \\
    Binary IDWT & $0.9831$ & $0.0711$ \\
    Model Without Content Loss & $0.9759$ & $0.0819$ \\
    Model Without Style Loss & $0.9736$ & $0.0916$ \\
    Sharp IDWT & $0.9680$ & $0.1005$ \\
    Blurry Test Set & $0.7565$ & $0.3407$ \\
\hline
\end{tabular}
\end{center}
\caption{Comparisons of different versions of the model by eliminating certain loss terms. Metrics obtained from VeriFinger SDK.}\label{ablation-table}
\end{table}

\section{Conclusion}
In summary, we proposed a fingerphoto deblurring architecture FDWST that utilizes techniques from both style transfer and the discrete wavelet transform. We demonstrated that our model not only improves fingerphoto quality, but also improves matching performance across multiple verifiers. Based on these findings, we assert that our Wavelet Style Transfer is proficient in transferring sharpness information and holds significant potential for various future applications.

{\small
\bibliographystyle{ieee}
\bibliography{egbib}
}

\end{document}